\documentclass{article}
\usepackage[preprint]{neurips_2025}

\usepackage[utf8]{inputenc} % allow utf-8 input
\usepackage[T1]{fontenc}    % use 8-bit T1 fonts
\usepackage{hyperref}       % hyperlinks
\usepackage{booktabs}       % professional-quality tables
\usepackage{nicefrac}       % compact symbols for 1/2, etc.
\usepackage{microtype}      % microtypography
\usepackage{xcolor}         % colors

\usepackage{mathtools}
\usepackage{amssymb}
\usepackage{amsthm}

\usepackage{wasysym}
\usepackage{annotate-equations}
\usepackage{algorithm}
\usepackage[indLines=false, spaceRequire=false]{algpseudocodex}
\newcommand\eg{\emph{e.g.,} }
\newcommand\ie{\emph{i.e.,} }
\newcommand\R{Replace}
\newcommand\A{Accumulate-Subsample}
\newcommand\TN{\textsf{TinyStories}}

\usepackage[capitalize, noabbrev]{cleveref}

\theoremstyle{plain}
\newtheorem{theorem}{Theorem}[section]
\newtheorem{proposition}[theorem]{Proposition}

\theoremstyle{definition}

\theoremstyle{remark}
\newtheorem{remark}[theorem]{Remark}

\title{Theoretical Proof that Auto-regressive Language Models Collapse when Real-world Data is a Finite Set}

\author{%
  Lecheng Wang$^{1,2}$, Xianjie Shi$^{1,2}$, Ge Li$^{1,2}$\footnotemark~, Jia Li \male$^{1,2}$, Xuanming Zhang$^3$,\\
  \textbf{Yihong Dong$^{1,2}$, Wenpin Jiao$^{1,2}$, Hong Mei$^{1,2}$}\\
  $^1$Key Laboratory of High Confidence Software Technologies (Peking University), Ministry of Education\\
  $^2$School of Computer Science, Peking University, Beijing, China\\
  $^3$University of Wisconsin-Madison\\
  \texttt{wanglecheng@stu.pku.edu.cn, lige@pku.edu.cn}\\
}

\begin{document}

\maketitle

{
  \renewcommand\thefootnote{\fnsymbol{footnote}}
  \footnotetext{Corresponding author}
}\setcounter{footnote}{0}

\begin{abstract}
  Auto-regressive language models (LMs) have been widely used to generate data in data-scarce domains to train new LMs, compensating for the scarcity of real-world data. Previous work experimentally found that LMs collapse when trained on recursively generated data. This paper presents a theoretical proof: once a corpus (such as a subset of the World Wide Web) begins to incorporate generated data and no new real-world data is added to the corpus, then no matter how small the amount of data each LM generates and contributes to the corpus, LM collapse is inevitable after sufficient time. This finding suggests that attempts to mitigate collapse by limiting the \emph{quantity} of synthetic data in the corpus are fundamentally insufficient. Instead, avoiding collapse hinges on ensuring the \emph{quality} of synthetic data.\footnote{The source code is available in the online data warehouse: \url{https://github.com/wanglc02/generated-data}}
\end{abstract}

\section{Introduction}
\label{sec: intro}

% 大背景，大问题
Auto-regressive language models (LMs)\footnote{In this paper, unless otherwise specified, LM refers to an \emph{auto-regressive} language model.} like ChatGPT \citep{ChatGPT} and Llama \citep{Llama} are widely used to generate text on the World Wide Web, such as articles and computer source code. Meanwhile, practitioners collect large amounts of text from the Web to train the next generations of LMs. Thus, the collected training corpus will inevitably include text generated by existing LMs. However, it is unclear whether the generated text in the training corpus has an inevitable negative impact on LMs \citep{Flaws}. Clarifying this issue is crucial for model trainers who rely on scraping a large amount of unvetted, potentially synthetic data from the Web to train LMs. It also critically applies to trainers using synthetic data from LMs to compensate for the scarcity of real-world data sources. This includes scenarios like using chain-of-thought data when training instruct models \citep{Self-Instruct, WizardLM}, or leveraging synthetic data in fields such as physics or computational biology when training corresponding large models \citep{Loong}.

% 小背景，小问题
A pioneer study \citep{Nature} warned that AI models, including LMs, can \emph{collapse} when trained on recursively generated data. Studies related to model collapse \citep{Feedback, MAD, Stability, Accumulate, Thrive} usually train the first-generation model with initial data, mix the data generated by the $(n-1)$-th generation model with previous data in different proportions to train the $n$-th generation model, and observe whether the evaluation performance of the model deteriorates, \ie the model \emph{collapses}. Although these studies revealed potentially adverse effects of generated data on many LMs, \eg OPT-125M \citep{OPT}, GPT, and Llama, it does not necessarily follow that all LMs will collapse if trained in the above-mentioned way. Therefore, a theoretical proof of LM collapse is essential.

% 我背景，我问题
Previous theoretical works proved that Gaussian model \citep{Nature}, linear regression model \citep{Regression, Accumulate}, simplified (non-autoregressive) LM \citep{COLM}, and toy bigram LLMs (large language models) \citep{Tails} can collapse. \citet{Strong} showed that even the smallest fraction of synthetic data (\eg as little as 1\% of the total training dataset) can still lead to the collapse of linear model and random projection model. However, to our knowledge, no one has proven that auto-regressive LMs will collapse. We believe that simply drawing analogies between LMs and other generative models cannot prove that LMs also conform to these collapse laws.

Recognizing this research gap, we present the first theoretical proof of collapse in auto-regressive LMs, demonstrating such collapse in data-scarce domains---specifically, when the real-world data constitutes a finite set. Our motivation is to hope that the theoretical analysis approach we have discovered can assist researchers in using theoretical tools to further explore the dynamics of LM collapse in domains with more complex data sources. In addition, we prove that LM collapse occurs regardless of the rate at which generated data infiltrates the initial real-world dataset in our scenario. This reveals that preventing LM collapse seems to rely more on filtering out low-quality data before training than on restricting generated data from entering the training corpus.

% Our experiments validate our proof and also reveal two findings related to LM applications: \ding{182} Training on recursively generated text inevitably reduces the LM's capabilities to generate text that is grammatically correct and consistent with human instructions. \ding{183} Training on recursively generated text significantly harms the performance of LMs on downstream tasks---that after finetuning on real data, they perform no better than a randomly initialized LM directly finetuned on real data. The experiments and findings are described in \cref{sec: experiments}.

\section{Proof: language model collapse}

\citet{Thrive} empirically observed that auto-regressive LMs collapse when recursively trained under the condition that real-world data constitutes a finite set. This scenario encompasses two data paradigms discussed in their paper: \emph\R~and \emph\A.

In the \emph\A~paradigm, a corpus progressively incorporates generated data, from which training samples are drawn for each LM. Our core contribution is a proof of LM collapse in this paradigm.

The \emph\R~paradigm is a classic paradigm first proposed by \citet{Nature}. In this paradigm, each LM can only access the data generated by the previous LM during training. Notably, the collapse of LMs in this paradigm has remained unproven until now.

As \citet{Thrive} noted, \emph\R~is an unrealistic model of reality: in practice, we do not discard real-world data after training our first LM. By contrast, \emph\A~reflects data accumulation (analogous to real-world Web data growth), where LMs are trained under a fixed computational budget---a scenario aligned with the resource constraints of individual practitioners. Given the similarity of proof strategies for both paradigms, we address them jointly. Our proof proceeds through the following steps:

\begin{itemize}
    \item \cref{sec: assumptions} formally defines the model of \emph{LM collapse} and establishes the foundational assumptions underpinning our Main Theorem.

    \item \cref{sec: main} formulates the Main Theorem in a self-contained manner.

    \item \cref{sec: 1,sec: n,sec: solve,sec: proof} present the theorem's proof through four phases:
    \begin{itemize}
        \item \cref{sec: 1} quantifies non-negative errors between the output distribution of generation 1 and the distribution of the initial corpus.
        \item \cref{sec: n} presents the errors from generation to generation.
        \item \cref{sec: solve} derives a closed-form expression for the output distribution of generation $n$ in terms of the distribution of the initial corpus and the errors introduced in each generation.
        \item \cref{sec: proof} demonstrates that these errors form positive series as $n \to \infty$, causing the output distribution of generation $n$ to approach arbitrarily closely a term composed solely of the errors introduced in each generation.
    \end{itemize}

    \item Finally, \cref{sec: conclusion} points out that the amount of generated data incorporated into the corpus in each generation does not affect the proof process. No matter how restricted the amount of data generated by each LM that flows into the corpus is, LM collapse is inevitable.
\end{itemize}

\subsection{Assumptions}
\label{sec: assumptions}

\begin{figure}
    \centering
    \includegraphics[width=0.5\linewidth]{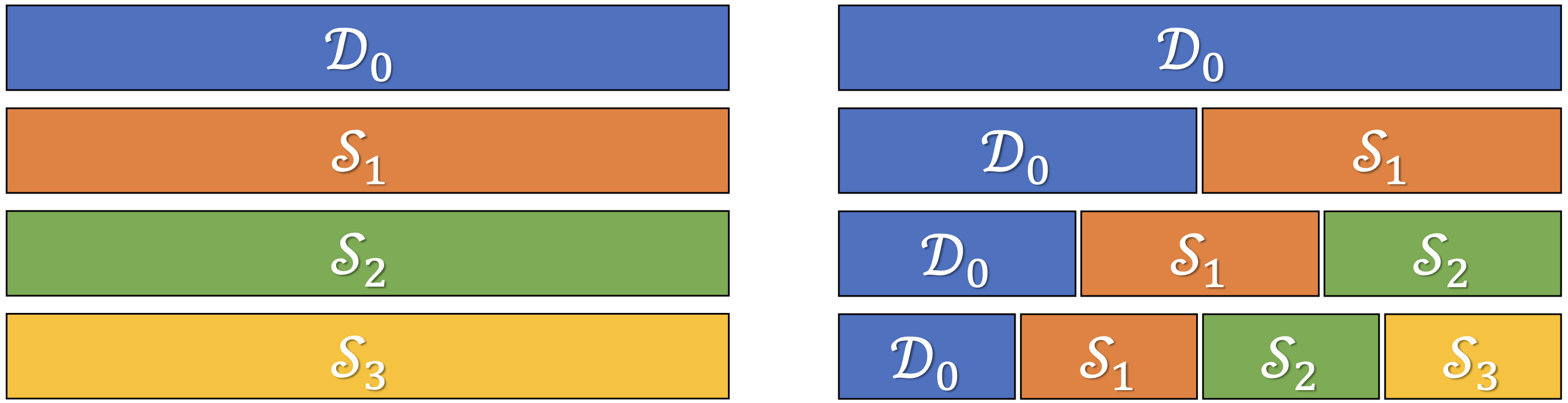}
    \caption{\textbf{Composition of the training set for each generation of language models (LMs) under different data paradigms.} On the left is \emph\R, and on the right is a special case of \emph\A~when $k = 1$. From top to bottom, each rectangle represents the composition of the training set of the 1st, 2nd, 3rd, and 4th generation LMs, respectively. $\mathcal D_0$, $\mathcal S_1$, $\mathcal S_2$, and $\mathcal S_3$ represent the initial corpus, the data generated by the 1st-generation LM, by the 2nd-generation LM, and by the 3rd-generation LM, respectively. The original figure is Figure 2 in the paper by \citet{Suffer}.}
    \label{fig: paradigms}
\end{figure}

Before presenting our Main Theorem, we make two assumptions. First, the two data paradigms used to train each LM involved in our theorem are defined as described in \cref{sec: def} and \cref{fig: paradigms}. Second, the meanings of all notations follow the specifications in \cref{sec: note}.

\subsubsection{Definitions}
\label{sec: def}

\paragraph\R\label{sec: replace} The first LM (generation 1) is trained on an initial corpus. The $n$-th LM (generation $n$) is trained on the data generated by the $(n-1)$-th LM for each $n \ge 2$.

\paragraph\A\label{sec: subsample} The first-generation LM is trained on an initial corpus. For each subsequent generation $n \ge 2$, the $n$-th LM is trained on the same total volume of data as generation 1. Of this data, a proportion $\frac 1 {1+(n-1)k}$ is sourced from the initial corpus. Meanwhile, each of the preceding $n-1$ LMs (from generation 1 to $n-1$) contributes a proportion $\frac k {1+(n-1)k}$ of the data, where $k > 0$ is a real-valued parameter representing the rate at which generated data is integrated into the corpus. This is because we assume that the data \emph{accumulate} in the following way: The data generated by generation 1 is added to the initial corpus, and the amount of data added to the corpus is $k$ times the initial data amount. If $k$ is equal to 1, then for each generation, the amount of generated data added to the corpus is as large as the initial data amount. This special case is the \emph{balanced} data cycle among the four data cycles defined by \citet{Suffer}. Subsequently, a subset as large as the initial corpus is \emph{subsampled}\footnote{Assume that the initial corpus is so large that the sampling error in the sampling process can be ignored. We will prove that even if the sampling error is ignored, the LMs will still collapse.} from the mixed corpus as the training set for generation 2. The data generated by generation 2 is added to the corpus, increasing the amount of data in the corpus by an amount equivalent to $k$ times the amount of data in the initial corpus. The training set for generation 3 is also \emph{subsampled} from the mixed corpus, and so on. In this way, the amount of data in the corpus grows, while the amount of training data for each generation of LMs is constant.

\subsubsection{Notations}
\label{sec: note}

\begin{itemize}
    \item $V$ is a finite set of tokens. Tokens are the basic units in LMs, often being words or meaningful subwords. Common tokens are collected as a vocabulary $V$.

    \item $\boldsymbol x \in V^*$ denotes a sequence of tokens.

    \item Vector $(p(v_1|\boldsymbol x),~p(v_2|\boldsymbol x),~\dots)$ is a probability distribution on the vocabulary $V$. It denotes the ground-truth distribution of the initial corpus. $\forall v_i \in V~~p(v_i|\boldsymbol x) \triangleq \frac {N_{\boldsymbol x;v_i}} {N_{\boldsymbol x}}$, where $N_{\boldsymbol x}$ denotes the number of occurrences of $\boldsymbol x$, $N_{\boldsymbol x;v_i}$ denotes the number of occurrences of the concatenation of $\boldsymbol x$ and $v_i$ in the corpus, subject to $N_{\boldsymbol x} = \sum_{v_i \in V} N_{\boldsymbol x;v_i}$.

    \item An LM takes an $\boldsymbol x$ (often called `context' or `prompt') as input.

    \item The LM at generation $n$ outputs a vector $(\hat p_n(v_1|\boldsymbol x),~\hat p_n(v_2|\boldsymbol x),~\dots)$, a probability distribution representing its prediction of the next token $v_i$ on the vocabulary $V$. \textbf{Assume that:}
    \begin{itemize}
        \item $\hat p_n(v_i|\boldsymbol x) \ne 0$. This assumption is plausible because the output probability of an LM is usually calculated by a logistic function whose range is $(0,1)$.
    \end{itemize}

    \item Non-negative quantities $\alpha_{\boldsymbol x}$ and $\alpha_{\boldsymbol x;v_i}$ (specific to $\boldsymbol x$) represent the discrepancies between $\hat p_1(v_i|\boldsymbol x)$ and the true probability $p(v_i|\boldsymbol x) = \frac {N_{\boldsymbol x;v_i}} {N_{\boldsymbol x}}$.

    \item When tracking the same specific $\boldsymbol x$ across generations, let non-negative quantities $\alpha[n]$ and $\alpha_i[n]$ denote the errors introduced during the training of the $n$-th generation LM. For clarity, we omit all subscripts $\boldsymbol x$ and simplify subscripts $\boldsymbol x;v_i$ to $i$. Specifically, $\alpha[1] \triangleq \alpha_{\boldsymbol x}$, $\alpha_i[1] \triangleq \alpha_{\boldsymbol x;v_i}$. \textbf{Assume that:}
    \begin{itemize}
        \item There exists an index $i$ for which $\alpha_i[n]$ does not tend to zero as $n \to \infty$. This assumption is made based on the following intuition: with $\alpha_i[n]$ modeling the errors, if the training hyperparameters are the same for every LMs, rarely the error factors will not exist in the later LMs, making $\alpha_i[n]$ tend to zero when $n \to \infty$.
    \end{itemize}
\end{itemize}

\subsection{Main Theorem}
\label{sec: main}

\begin{theorem}[LM collapse]
For any specific token sequence $\boldsymbol x$ and token $v_i$, along with corresponding $\alpha[j]$ and $\alpha_i[j]$, under the \R~paradigm,
\[
\forall \epsilon > 0~\exists n_0 \in \mathbb N~\forall n > n_0~\left(\left|\hat p_n(v_i|\boldsymbol x) - \frac {\sum_{j=1}^n \alpha_i[j]} {\sum_{j=1}^n \alpha[j]}\right| < \epsilon\right).
\]
Under the \A~paradigm,
\[
\forall \epsilon > 0~\exists n_1 \in \mathbb N~\forall n > n_1~\left(\left|\hat p_n(v_i|\boldsymbol x) - \frac {k\sum_{j=1}^{n-1} \frac 1 {1+jk} \alpha_i[j]} {k\sum_{j=1}^{n-1} \frac 1 {1+jk} \alpha[j]}\right| < \epsilon\right),
\]
where $k$, as defined in \cref{sec: subsample}, is the rate at which generated data is added to the corpus.
\end{theorem}

\begin{remark}
This theorem indicates that when $n$ is sufficiently large, $\hat p_n(v_i|\boldsymbol x)$ can approach arbitrarily closely a term composed solely of the errors $\alpha[n]$ and $\alpha_i[n]$ introduced in each generation, rather than necessarily being close to the ground-truth distribution of the initial corpus. This may cause the cross-entropy between $\hat p_n(v_i|\boldsymbol x)$ and $p(v_i|\boldsymbol x)$ to increase, leading to an increase in the test loss of the LM on the initial corpus. This is consistent with the concept of model collapse.\footnote{Prior theoretical works on model collapse \citep{Accumulate, Thrive, Strong} often proved that a generative model's test loss tends to infinity as the number of generations grows. However, LMs have their special property---their test loss has an upper bound. This is because there is an upper bound on the cross-entropy between any two probability distributions on the vocabulary (here, the distribution predicted by the LM and the ground-truth distribution). Thus, the test loss of LMs does not tend to infinity as the number of generations grows. Therefore, we turn to the proof objective used here.}
\end{remark}

To prove the Main Theorem, we will derive the expression for $\hat p_n(v_i|\boldsymbol x)$. \cref{sec: 1} will present the expression for the initial value $\hat p_1(v_i|\boldsymbol x)$, \cref{sec: n} will present the recurrence relation that relates $\hat p_n(v_i|\boldsymbol x)$ to $\hat p_{n-1}(v_i|\boldsymbol x)$, and \cref{sec: solve} will solve the recurrence equation to obtain the expression for $\hat p_n(v_i|\boldsymbol x)$. In \cref{sec: proof}, we will examine the expression for $\hat p_n(v_i|\boldsymbol x)$ and prove that it satisfies the conditions stated in the Main Theorem.

\subsection{Generation 1: initial output distribution}
\label{sec: 1}

The training approaches for generation 1 under both data paradigms are identical, hence the following content in this \lcnamecref{sec: 1} applies to both paradigms: Regardless of what value $\hat p_1(v_i|\boldsymbol x)$ takes, it can be expressed through the combination of $p(v_i|\boldsymbol x)$ with non-negative errors---as per the conventions in \cref{sec: note}, we denote these non-negative quantities using $\alpha_{\boldsymbol x}$ and $\alpha_{\boldsymbol x;v_i}$.

\begin{proposition}\label{proposition: initial}
$\forall \boldsymbol x \in V^*~\forall v_i \in V,~~~\hat p_1(v_i|\boldsymbol x),~p(v_i|\boldsymbol x) = \frac {N_{\boldsymbol x;v_i}} {N_{\boldsymbol x}},~\alpha_{\boldsymbol x}\text{, and }\alpha_{\boldsymbol x;v_i}$ are related by
\begin{equation}\label{first}
\hat p_1(v_i|\boldsymbol x) = \frac {N_{\boldsymbol x;v_i} + \alpha_{\boldsymbol x;v_i}} {N_{\boldsymbol x} + \alpha_{\boldsymbol x}},
\end{equation}
where
\[
\alpha_{\boldsymbol x} \triangleq \sum_{v_i \in V} \alpha_{\boldsymbol x;v_i}.
\]
\end{proposition}

\begin{proof}See \cref{sec: p-proof}.\end{proof}

\subsection{Effects of per-generation errors on the output distribution of generation \texorpdfstring{$n~(n \ge 2)$}{n (n >= 2)}}
\label{sec: n}

Subsequent LMs are trained on recursively generated data, with the specific formulation depending on the data paradigm employed. We analyze the two paradigms separately below.

\subsubsection\R

\begin{proposition}\label{proposition: replace}
Under the \R~paradigm, given $\hat p_{n-1}(v_i|\boldsymbol x) = \frac {y_i[n-1]} {y[n-1]}$ where $y[n-1]$ and $y_i[n-1]$ are positive scalars, $\hat p_n(v_i|\boldsymbol x)$ admits a recursive formulation through non-negative quantities $\alpha[n]$ and $\alpha_i[n]$:
\begin{equation}\label{replace-subsequent}
\hat p_n(v_i|\boldsymbol x) = \frac {y_i[n-1] + \alpha_i[n]} {y[n-1] + \alpha[n]},
\end{equation}
where
\[
\alpha[n] \triangleq \sum_{v_i \in V} \alpha_i[n].
\]
\end{proposition}

\begin{proof}See \cref{sec: r-proof}.\end{proof}

\subsubsection\A

Similarly, following the definition of training set composition for each generation within the \emph\A~paradigm, we formulate the following proposition:

\begin{proposition}
Under the \A~paradigm defined in \cref{sec: subsample} with accumulation rate $k$, given $\forall n~\hat p_n(v_i|\boldsymbol x) = \frac {y_i[n]} {y[n]}$ where $y[n]$ and $y_i[n]$ are positive scalars, $\hat p_n(v_i|\boldsymbol x)$ admits a recursive formulation
\begin{equation}\label{subsample-subsequent}
\hat p_n(v_i|\boldsymbol x) = \frac {\frac {N_{\boldsymbol x;v_i} + k\sum_{j=1}^{n-1} y_i[j]} {1+(n-1)k} + \alpha_i[n]} {\frac {N_{\boldsymbol x} + k\sum_{j=1}^{n-1} y[j]} {1+(n-1)k} + \alpha[n]},
\end{equation}
where
\[
\alpha[n] \triangleq \sum_{v_i \in V} \alpha_i[n].
\]
\end{proposition}

\subsection{Solving for the output distribution of generation \texorpdfstring{$n$}{n}}
\label{sec: solve}

\subsubsection\R

\cref{first,replace-subsequent} respectively present the initial value $\hat p_1(v_i|\boldsymbol x)$ and the equation that relates $\hat p_n(v_i|\boldsymbol x)$ to its preceding values. Combining both equations yields the following recurrence equation
\begin{equation}\label{replace-equation}
\hat p_n(v_i|\boldsymbol x) =
\begin{cases}
\displaystyle \frac {N_{\boldsymbol x;v_i} + \alpha_i[1]} {N_{\boldsymbol x} + \alpha[1]},&n = 1,\\
\\
\displaystyle \frac {y_i[n-1] + \alpha_i[n]} {y[n-1] + \alpha[n]},&n \ge 2.\\
\end{cases}
\end{equation}
A solution is
\[
\begin{cases}
y_i[n] &=~~N_{\boldsymbol x;v_i} + \sum_{j=1}^n \alpha_i[j]\\
y[n] &=~~N_{\boldsymbol x} + \sum_{j=1}^n \alpha[j]\\
\end{cases},
\]
\begin{equation}\label{replace-solution}
\hat p_n(v_i|\boldsymbol x) = \frac {y_i[n]} {y[n]} = \frac {N_{\boldsymbol x;v_i} + \sum_{j=1}^n \alpha_i[j]} {N_{\boldsymbol x} + \sum_{j=1}^n \alpha[j]}.
\end{equation}
Finding this solution is quite straightforward. We provide a proof that this is a solution to the equation in \cref{sec: replace-proof}.

\subsubsection\A

Combining \cref{first,subsample-subsequent} yields
\begin{equation}\label{subsample-equation}
\hat p_n(v_i|\boldsymbol x) =
\begin{cases}
\displaystyle \frac {N_{\boldsymbol x;v_i} + \alpha_i[1]} {N_{\boldsymbol x} + \alpha[1]},&n = 1,\\
\\
\displaystyle \frac {\frac {N_{\boldsymbol x;v_i} + k\sum_{j=1}^{n-1} y_i[j]} {1+(n-1)k} + \alpha_i[n]} {\frac {N_{\boldsymbol x} + k\sum_{j=1}^{n-1} y[j]} {1+(n-1)k} + \alpha[n]},&n \ge 2.\\
\end{cases}
\end{equation}
A solution is
\[
\begin{cases}
y_i[n] &=~~N_{\boldsymbol x;v_i} + \alpha_i[n] + k\sum_{j=1}^{n-1} \frac 1 {1+jk} \alpha_i[j]\\
y[n] &=~~N_{\boldsymbol x} + \alpha[n] + k\sum_{j=1}^{n-1} \frac 1 {1+jk} \alpha[j]\\
\end{cases},
\]
\begin{equation}\label{subsample-solution}
\hat p_n(v_i|\boldsymbol x) = \frac {y_i[n]} {y[n]} = \frac {N_{\boldsymbol x;v_i} + \alpha_i[n] + k\sum_{j=1}^{n-1} \frac 1 {1+jk} \alpha_i[j]} {N_{\boldsymbol x} + \alpha[n] + k\sum_{j=1}^{n-1} \frac 1 {1+jk} \alpha[j]}.
\end{equation}
This solution is not obvious. We provide a walkthrough of how to find this solution in \cref{sec: walkthrough}. We prove that this is a solution to the equation in \cref{sec: subsample-proof}.

\subsection{Completion of the Main Theorem proof}
\label{sec: proof}

Putting \cref{replace-solution,subsample-solution} here again, we can see that, under both paradigms, the errors (denoted by non-negative numbers $\alpha[n]$ and $\alpha_i[n]$) accumulate as $n$ increases.

Specifically, under the \emph\R~paradigm,
\vspace{2em}
\[
\hat p_n(v_i|\boldsymbol x) = \frac {\eqnmarkbox[blue]{ni}{N_{\boldsymbol x;v_i}} + \eqnmarkbox[red]{ai}{\sum_{j=1}^n \alpha_i[j]}} {\eqnmarkbox[blue]{n}{N_{\boldsymbol x}} + \eqnmarkbox[red]{a}{\sum_{j=1}^n \alpha[j]}}.
\]
\annotate[yshift=1em]{left}{ai, a}{positive series when $n \to \infty$}
\annotate[yshift=-2.5em]{below, left}{ni, n}{constants}
\vspace{1em}

Under the \emph\A~paradigm,
\vspace{2em}
\[
\hat p_n(v_i|\boldsymbol x) = \frac {\eqnmarkbox[blue]{ni}{N_{\boldsymbol x;v_i}} + \alpha_i[n] + \eqnmarkbox[red]{ai}{k\sum_{j=1}^{n-1} \frac 1 {1+jk} \alpha_i[j]}} {\eqnmarkbox[blue]{n}{N_{\boldsymbol x}} + \alpha[n] + \eqnmarkbox[red]{a}{k\sum_{j=1}^{n-1} \frac 1 {1+jk} \alpha[j]}}.
\]
\annotate[yshift=1em]{left}{ai, a}{positive series when $n \to \infty$}
\annotate[yshift=-2.5em]{below, left}{ni, n}{constants}
\vspace{1em}

As annotated above, under both paradigms, terms related to errors become positive series when $n \to \infty$, while terms related to the distribution of the initial corpus, $N_{\boldsymbol x;v_i}$ and $N_{\boldsymbol x}$, are constants.

Taking the \emph\R~paradigm as an example: Under the assumption that there exists an index $i$ for which $\alpha_i[n]$ does not tend to zero as $n \to \infty$, it follows from the properties of positive series that the series $\sum_{j=1}^n \alpha[j]$ diverges. Consequently, for sufficiently large $n$, the difference between $\hat p_n(v_i|\boldsymbol x)$ and the ratio $\frac {\sum_{j=1}^n \alpha_i[j]} {\sum_{j=1}^n \alpha[j]}$ becomes arbitrarily small. Formally, this can be expressed as:
\[
\forall \epsilon > 0~\exists n_0 \in \mathbb N~\forall n > n_0~\left(\left|\hat p_n(v_i|\boldsymbol x) - \frac {\sum_{j=1}^n \alpha_i[j]} {\sum_{j=1}^n \alpha[j]}\right| < \epsilon\right).
\]
This proves the Main Theorem.

\begin{remark}
Under our assumptions, the distribution of the LM's prediction of the next token may not necessarily converge to a fixed distribution, since $\frac {\sum_{j=1}^n \alpha_i[j]} {\sum_{j=1}^n \alpha[j]}$ is not a constant but rather a function of $n$. Specifically, $\hat p_n(v_i|\boldsymbol x)$ may continue to vary with increasing $n$, regardless of how large $n$ becomes. Nonetheless, \textbf{we prove that when $n$ is sufficiently large, the output distribution of generation $n$ does not necessarily approximate the ground-truth distribution of the initial corpus---a property that holds irrespective of whether the output distribution converges to any particular distribution. \cref{sec: results} depicts this property.}
% In essence, the LM's prediction of the next token ceases to accurately reflect the statistical properties inherent in the original real-world data distribution.
\end{remark}

Under the \emph\A~paradigm and the same assumptions, the series $k\sum_{j=1}^{n-1} \frac 1 {1+jk} \alpha[j]$ also diverges. This leads to the following conclusion in the Main Theorem:
\[
\forall \epsilon > 0~\exists n_1 \in \mathbb N~\forall n > n_1~\left(\left|\hat p_n(v_i|\boldsymbol x) - \frac {k\sum_{j=1}^{n-1} \frac 1 {1+jk} \alpha_i[j]} {k\sum_{j=1}^{n-1} \frac 1 {1+jk} \alpha[j]}\right| < \epsilon\right).
\]
The following \lcnamecref{sec: conclusion} will provide some final discussion regarding this conclusion for the \emph\A~paradigm.

\subsection{Conclusion on the existence and implications of LM collapse}
\label{sec: conclusion}

\cref{sec: subsample} defines $k$ as the rate at which the generated data enters the corpus. Note that the magnitude of $k$ does not affect our proof. This property indicates that once the corpus begins to expand due to generated data, while the amount of training data for each generation of LMs remains unchanged, LM collapse will eventually occur. It is just a matter of time. In practice, if the corpus is large and $k$ is small, we can be relatively sure that the collapse of LM will not occur in the short term. However, by proving the \emph{existence} of LM collapse, we reasonably express our concerns about the current situation where an increasing amount of generated data is used in LM training.

\section{Experiments}
\label{sec: experiments}

We empirically analyze the output distribution of each LM generation to demonstrate our theoretical findings.

\subsection{Experimental setup}
\label{sec: setup}

Following the \emph\R~data paradigm defined in \cref{sec: replace}, we train generation $n$ exclusively on the data generated by generation $n-1$. As demonstrated by our Main Theorem, the outcomes of LM collapse exhibit no fundamental distinction between the \emph\R~and \emph\A~paradigms as $n$ approaches infinity. Since LM collapse may take many generations, using \emph\R~allows for earlier observation of experimental phenomena.

\subsubsection{Training corpora}

For generation 1, we select a popular corpus named \TN~\citep{TinyStories} as its training corpus. \TN~consists of $2.12 \times 10^6$ short English stories. The stories contain only words that most 3 to 4 year olds would typically understand. They are so simple that they can be used to train and evaluate small language models (SLMs) that are much smaller than the state-of-the-art models, yet still enable them to produce a diverse set of fluent and consistent stories. The training of SLMs on \TN~can typically be done in less than a day on a single GPU from scratch. Therefore, corpora like \TN~facilitate our experiments.

For generation $n~(n \ge 2)$, we use generation $n-1$ to generate stories as the training corpus. Each story is generated freely, \ie starting from an empty prompt. The tokens of the story are repeatedly sampled from the output distribution of the LM. The scale of the generated corpus is the same as \TN. Details can be found in \cref{sec: recursive}, and \cref{tab: excerpts} in which segment of the corpora generated by LMs of various generations are shown, demonstrating what these corpora look like.

\subsubsection{Language models}

Our LMs utilize the classic GPT-Neo architecture \citep{GPT-Neo}. We train two series of LMs. In one series, all LMs have 1 million (1M) parameters. In the other series, all LMs have 33 million (33M) parameters. For detailed architectures, please refer to \cref{sec: architecture}. We train the 1M LMs on the text generated by the previous 1M LMs and the 33M LMs on the text generated by the previous 33M LMs. One can inspect if LMs with more parameters are more resilient to performance degradation during recursive training.

\subsubsection{Inspections carried out on each generation of LMs}

\paragraph{Output distribution of generation $n$ on fixed prompt (\cref{fig: output-33M})} We demonstrate the numerical values of $\hat p_n(v_i|\boldsymbol x)$ in our theoretical part by inputting the same input $\boldsymbol x$ into each generation of LMs.

\paragraph{Corpus-level divergence of generation $n$ (\cref{fig: perplexity})} We compare the `average distance' between the output distributions of each generation and the distributions of the initial corpus using model perplexity (test loss) on the validation set of \TN. The formula for calculating this distance can be found in \cref{sec: perplexity}.

\subsection{Experimental results}
\label{sec: results}

\begin{figure}[h]
    \centering
    \includegraphics[width=0.6\linewidth]{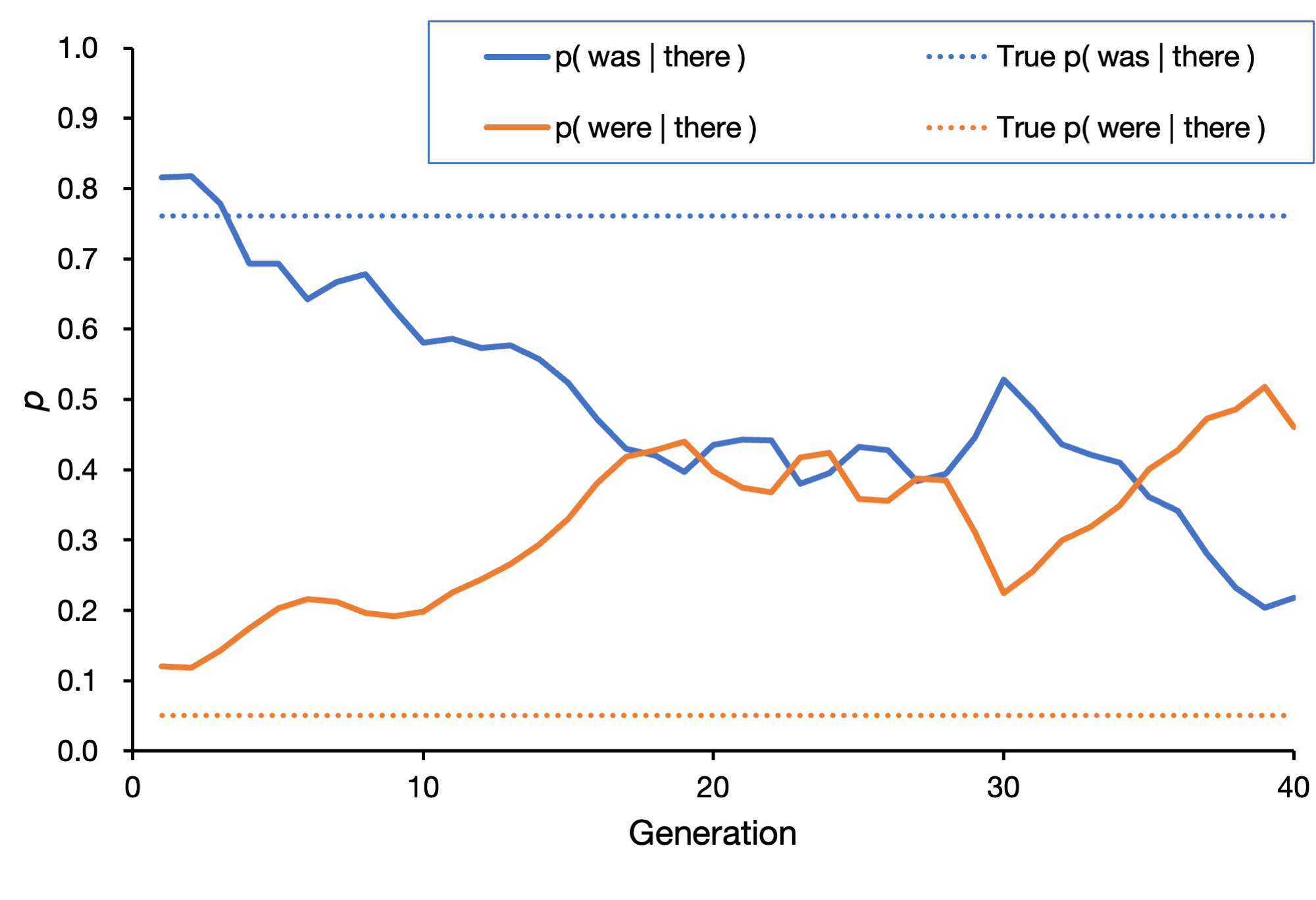}
    \caption{\textbf{Language model (LM) collapse is a degenerative process whereby, over generations, LMs forget the underlying text distribution of the initial training corpus.} Our experiment begins with an initial corpus used to train the LM at generation 1. Then, generation 2 is trained using the text generated by generation 1, generation 3 using the text generated by generation 2, and so on. The \lcnamecref{fig: output-33M} depicts the process of LM collapse. The dotted lines True $p(\text{was}|\text{there})$ and True $p(\text{were}|\text{there})$ refer to the probabilities of two phrases `there was' and `there were' given `there' in the initial corpus \TN, which are 0.76 and 0.05 respectively. The first-generation LM with 33 million parameters trained on it can learn this probability well (the solid lines $p(\text{was}|\text{there})$ and $p(\text{were}|\text{there})$ represent the probabilities that the LMs predicts the next token after `there' to be `was' and `were' respectively). However, as $n$ increases, the solid lines gradually deviate from the dotted lines, underscoring the growing disconnect between the output of LMs and the initial training text.}
    \label{fig: output-33M}
\end{figure}

\cref{fig: output-33M} demonstrates the evolution of $\hat p_n(v_i|\boldsymbol x)$ values for $n$ ranging from 1 to 40, where the token sequence $\boldsymbol x$ is a single token `there' and $v_i$ represents either `was' or `were'. This case study was specifically selected for its clear demonstration of LM prediction divergence from the ground-truth distribution across generations. In the initial \TN~corpus, we observe $p(\text{was}|\text{there}) = 0.76$ and $p(\text{were}|\text{there}) = 0.05$, represented by horizontal reference lines of True $p(\text{was}|\text{there})$ and True $p(\text{were}|\text{there})$ in the \lcnamecref{fig: output-33M}.\footnote{Most of the stories are in the past tense, so the probabilities of `there was' and `there were' occurring are relatively high, while the probabilities of `there is' and `there are' occurring are relatively low.}

\begin{figure}
    \centering
    \includegraphics[width=0.5\linewidth]{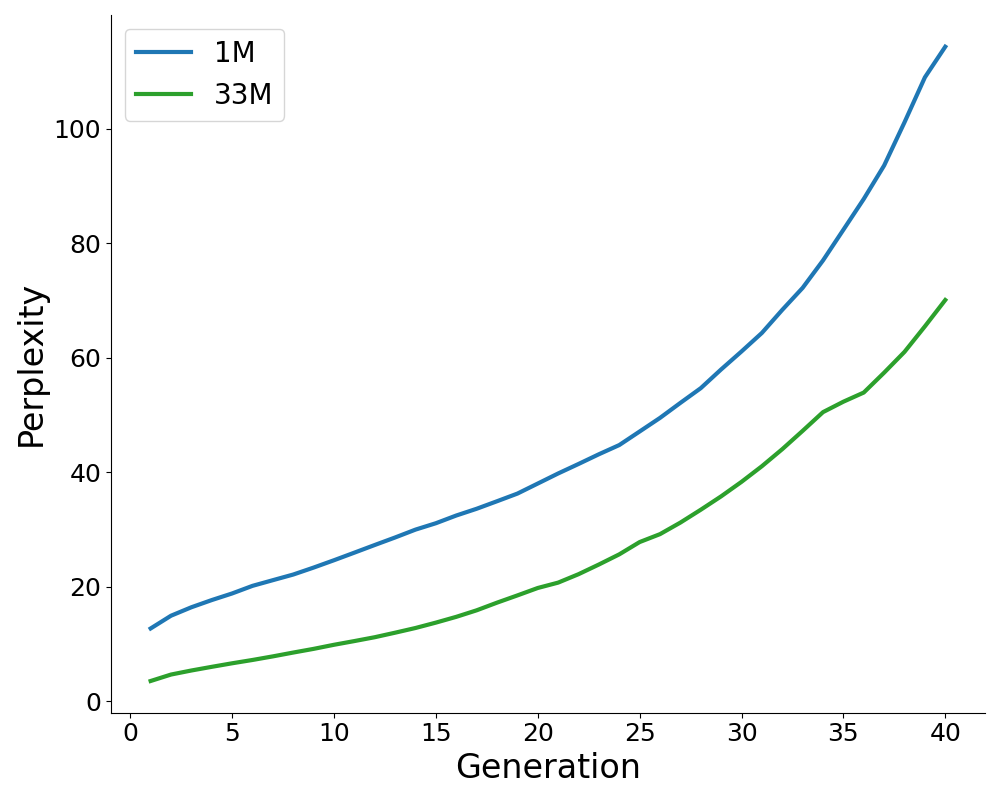}
    \caption{\textbf{Perplexity of LMs with 1M and 33M parameters over 40 generations evaluated on the validation set of the \TN~dataset.} Model perplexity is not only an indicator of the performance of LMs, but also a measure of the `distance' between the distributions learned by the LM and the distributions of the initial corpus. (The formula for calculating perplexity can be found in \cref{sec: perplexity}.) The increase in perplexity is due to the increase in this distance, which means that the learned distribution is becoming increasingly deviated from the initial corpus. This corroborates our theoretical result.}
    \label{fig: perplexity}
\end{figure}

Our analysis reveals that LM outputs initially approximate the corpus distribution well, with $p_n(\text{was}|\text{there})$ and $p_n(\text{were}|\text{there})$ closely matching their respective ground-truth probabilities when $n$ is small, indicating that our LMs are able to learn this distribution in the initial corpus. However, as $n$ increases, the probabilities output by LMs deviate significantly from those in the initial corpus, which corroborates the result of our theoretical proof. Note that this deviation does not increase monotonically with $n$. For example, the difference between $p_n(\text{was}|\text{there})$ and $p(\text{was}|\text{there})$ temporarily decreases in some generations. It does not violate our theory because even if the probability fluctuates as $n$ increases and continues to fluctuate forever, we have proven that the LM's prediction of the next token ceases to accurately reflect the original real-world data distribution. Moreover, from a statistical perspective, the perplexity (equivalent to test loss) of generation $n$ on the initial corpus can be used to measure the degree to which its output distribution deviates from the ground-truth distribution on average. As shown in \cref{fig: perplexity}, this metric reveals a clear trend of increasing deviation when local fluctuations from individual token examples are aggregated, providing stronger evidence of LM collapse than single-token trajectories. We make available our code for reproducing our results as well as probing output probabilities with custom prompts $\boldsymbol x$ in the supplementary material.

\section{Discussion}

Our theoretical proof establishes that LM collapse occurs regardless of the rate $k$ at which generated data infiltrates the initial real-world dataset, as long as no new real-world data is introduced. This finding suggests that attempts to mitigate collapse by limiting the \emph{quantity} of synthetic data in the corpus are fundamentally insufficient. Instead, avoiding collapse hinges on ensuring the \emph{quality} of synthetic data.

This conclusion aligns with contemporaneous works exploring collapse prevention. For example, work by \citet{Synthesize} demonstrates that token-level editing techniques can effectively counteract collapse by refining synthetic data quality, even when models are trained recursively on edited outputs. Such approaches validate our theoretical insight: collapse arises not from the mere presence of synthetic data, but from its progressive divergence from the ground-truth data distribution across generations. By theoretically demonstrating that \emph{any} amount of uncurated synthetic data accumulation leads to eventual collapse, our work clarifies the futility of strategies focused solely on regulating the \emph{proportion} of generated data.

These findings call for research attention towards a critical direction: developing robust methods to assess and enhance the quality of synthetic data at scale. Promising avenues include (1) refining data filtering mechanisms to retain high-entropy, diverse patterns in generated text, (2) integrating human or automated feedback loops to improve synthetic data iteratively, and (3) hybrid training paradigms that strategically blend real and high-quality synthetic data. Ultimately, our proof underscores that the sustainability of LM training hinges not on avoiding synthetic data entirely, but on rigorously ensuring its fidelity to the original data distribution.

\section{Limitations}
\label{sec: lim}

We assume that real-world data is a finite set. This reflects the scenario of data-scarce domains. Outside this domain, when the proportion of real-world data in the training corpus does not approach zero as the generation number $n$ increases, LMs may still collapse. However, their output probabilities will not approach arbitrarily closely a term composed solely of the errors introduced in each generation, as stated in our Main Theorem.

Our work does not account for post-processing techniques applied to generated data. It also does not consider commonly used preprocessing methods for filtering data before training new LMs. Future research could address these limitations by relaxing our assumptions to accommodate more scenarios.

\bibliography{b}
\bibliographystyle{icml2025}

%%%%%%%%%%%%%%%%%%%%%%%%%%%%%%%%%%%%%%%%%%%%%%%%%%%%%%%%%%%%

\appendix

\section{Related work}

\citet{Optimize} observes that emerging theoretical analyses of self-consuming generative models (also known as iterative retraining) suggest model collapse or stability may occur depending on the proportion of generated data used in each retraining iteration. This does not contradict our theoretical result, as it addresses a fundamentally different experimental scenario. Upon examination, the cited stability result originates from \citet{Stability}. While their framework demonstrates non-collapsing models, our hypothesized scenario leads to model collapse. The key distinction lies in the assumptions:

\citet{Stability} postulates a persistent clean dataset that remains uncontaminated by generated content. Subsequent models are trained on hybrid datasets combining this pristine source with outputs from prior models, examining how incremental additions of synthetic data affect model behavior. Their analysis shows stability can be maintained when synthetic data proportions remain limited.

Conversely, our study investigates a distinct paradigm where generated content progressively contaminates a single evolving dataset, as illustrated in \cref{fig: paradigms}. In this irreversible contamination process, the original real-world data becomes asymptotically negligible through successive retraining cycles. Our research focuses specifically on this complete corpus pollution scenario, making it complementary to the work by \citet{Stability} while addressing different dimensions of the data contamination problem.

\section{Mathematical derivations}

\subsection{Proof of \cref{proposition: initial}}
\label{sec: p-proof}

Declare a real number $S$. Let
\[
\forall v_i \quad \alpha_{\boldsymbol x;v_i} \triangleq \hat p_1(v_i|\boldsymbol x) \cdot S + \hat p_1(v_i|\boldsymbol x) \cdot N_{\boldsymbol x} - N_{\boldsymbol x;v_i},
\]
then
\[
S \ge \frac {N_{\boldsymbol x;v_i} - \hat p_1(v_i|\boldsymbol x) \cdot N_{\boldsymbol x}} {\hat p_1(v_i|\boldsymbol x)} \implies \alpha_{\boldsymbol x;v_i} \ge 0,
\]
\[
S \ge \max_{v_i \in V} \frac {N_{\boldsymbol x;v_i} - \hat p_1(v_i|\boldsymbol x) \cdot N_{\boldsymbol x}} {\hat p_1(v_i|\boldsymbol x)} \implies \forall v_i~~\alpha_{\boldsymbol x;v_i} \ge 0.
\]
Notice that,
\[
\begin{split}
\alpha_{\boldsymbol x} &= \sum_{v_i \in V} \alpha_{\boldsymbol x;v_i}\\
&= \sum_{v_i \in V} (\hat p_1(v_i|\boldsymbol x) \cdot S + \hat p_1(v_i|\boldsymbol x) \cdot N_{\boldsymbol x} - N_{\boldsymbol x;v_i})\\
&= S \cdot \underbrace{\sum_{v_i \in V} \hat p_1(v_i|\boldsymbol x)}_{=1} ~+~ N_{\boldsymbol x} \cdot \underbrace{\sum_{v_i \in V} \hat p_1(v_i|\boldsymbol x)}_{=1} ~-~ \sum_{v_i \in V} N_{\boldsymbol x;v_i}\\
&= S + N_{\boldsymbol x} - N_{\boldsymbol x} \quad (\because \sum_{v_i \in V} N_{\boldsymbol x;v_i} = N_{\boldsymbol x})\\
&= S.\\
\end{split}
\]
Thus, $\forall v_i~\alpha_{\boldsymbol x;v_i} = \hat p_1(v_i|\boldsymbol x) \cdot \alpha_{\boldsymbol x} + \hat p_1(v_i|\boldsymbol x) \cdot N_{\boldsymbol x} - N_{\boldsymbol x;v_i} \ge 0$, which further implies
\[
\hat p_1(v_i|\boldsymbol x) = \frac {N_{\boldsymbol x;v_i} + \alpha_{\boldsymbol x;v_i}} {N_{\boldsymbol x} + \alpha_{\boldsymbol x}},
\]
and $\alpha_{\boldsymbol x;v_i}$ is non-negative.

% We can always find non-negative numbers
\begin{remark}
For any particular $\boldsymbol x$,
\[
\alpha_{\boldsymbol x}^\star \triangleq \max_{v_i \in V} \frac {N_{\boldsymbol x;v_i} - \hat p_1(v_i|\boldsymbol x) \cdot N_{\boldsymbol x}} {\hat p_1(v_i|\boldsymbol x)},
\]
is the minimum $\alpha_{\boldsymbol x}$ that satisfies the above proposition. $\alpha_{\boldsymbol x}^\star = 0$ if and only if the output distribution of the LM is the same as the distribution of the training corpus given $\boldsymbol x$, that is to say, there is no error, and
\[
\forall v_i \quad \hat p_1(v_i|\boldsymbol x) = \frac {N_{\boldsymbol x;v_i}} {N_{\boldsymbol x}}.
\]
Furthermore, in this situation, $\forall v_i~\alpha_{\boldsymbol x;v_i}^\star = 0$. \textbf{This property further illustrates that we can use non-negative numbers to model the errors between the output distribution of an LM and the distribution of its training corpus.}
\end{remark}

\subsection{Proof of \cref{proposition: replace}}
\label{sec: r-proof}

The existence proof follows identical reasoning to \cref{proposition: initial} through parameter substitution:
\[
\alpha_i[n] = \hat p_n(v_i|\boldsymbol x) \cdot S + \hat p_n(v_i|\boldsymbol x) \cdot y[n-1] - y_i[n-1],
\]
where
\[
S \ge \max_{v_i \in V} \frac {y_i[n-1] - \hat p_n(v_i|\boldsymbol x) \cdot y[n-1]} {\hat p_n(v_i|\boldsymbol x)}
\]
guarantees non-negativity.

\subsection{Proof that \cref{replace-solution} is a solution to the recurrence \cref{replace-equation}}
\label{sec: replace-proof}

Substitute $n = 1$ into
\[
\begin{cases}
y_i[n] &=~~N_{\boldsymbol x;v_i} + \sum_{j=1}^n \alpha_i[j]\\
y[n] &=~~N_{\boldsymbol x} + \sum_{j=1}^n \alpha[j]\\
\end{cases},
\]
we get
\[
\begin{cases}
y_i[1] &=~~N_{\boldsymbol x;v_i} + \alpha_i[1]\\
y[1] &=~~N_{\boldsymbol x} + \alpha[1]\\
\end{cases},
\]
which satisfies the initial condition
\[
\hat p_1(v_i|\boldsymbol x) = \frac {y_i[1]} {y[1]} = \frac {N_{\boldsymbol x;v_i} + \alpha_i[1]} {N_{\boldsymbol x} + \alpha[1]}.
\]
Substitute
\[
\begin{cases}
y_i[n-1] &=~~N_{\boldsymbol x;v_i} + \sum_{j=1}^{n-1} \alpha_i[j]\\
y[n-1] &=~~N_{\boldsymbol x} + \sum_{j=1}^{n-1} \alpha[j]\\
\end{cases}
\]
into the right hand side of
\[
\hat p_n(v_i|\boldsymbol x) = \frac {y_i[n-1] + \alpha_i[n]} {y[n-1] + \alpha[n]},
\]
we get
\[
\frac {y_i[n-1] + \alpha_i[n]} {y[n-1] + \alpha[n]} = \frac {N_{\boldsymbol x;v_i} + \sum_{j=1}^{n-1} \alpha_i[j] + \alpha_i[n]} {N_{\boldsymbol x} + \sum_{j=1}^{n-1} \alpha[j] + \alpha[n]} = \frac {N_{\boldsymbol x;v_i} + \sum_{j=1}^n \alpha_i[j]} {N_{\boldsymbol x} + \sum_{j=1}^n \alpha[j]} = \frac {y_i[n]} {y[n]} = \hat p_n(v_i|\boldsymbol x).
\]
This shows that the obtained solution satisfies the original recurrence equation.

\subsection{The thinking process of solving the recurrence \cref{subsample-equation}}
\label{sec: walkthrough}

We conjecture that \cref{subsample-equation} has a solution satisfying
\begin{equation}\label{n}
y_i[n] = \frac {N_{\boldsymbol x;v_i} + k\sum_{j=1}^{n-1} y_i[j]} {1+(n-1)k} + \alpha_i[n]
\end{equation}
Then $y_i[n-1]$ will satisfy
\begin{equation}\label{n-1}
y_i[n-1] = \frac {N_{\boldsymbol x;v_i} + k\sum_{j=1}^{n-2} y_i[j]} {1+(n-2)k} + \alpha_i[n-1]
\end{equation}
Multiply both sides of \cref{n} by $1+(n-1)k$ and both sides of \cref{n-1} by $1+(n-2)k$, we get
\[
(1+(n-1)k)y_i[n] = N_{\boldsymbol x;v_i} + k\sum_{j=1}^{n-1} y_i[j] + (1+(n-1)k)\alpha_i[n]
\]
\[
(1+(n-2)k)y_i[n-1] = N_{\boldsymbol x;v_i} + k\sum_{j=1}^{n-2} y_i[j] + (1+(n-2)k)\alpha_i[n-1]
\]
Subtract the two equations, we obtain
\[
(1+(n-1)k)y_i[n] - (1+(n-2)k)y_i[n-1] = ky_i[n-1] + (1+(n-1)k)\alpha_i[n] - (1+(n-2)k)\alpha_i[n-1]
\]
Add $(1+(n-2)k)y_i[n-1]$ to both sides, we have
\[
(1+(n-1)k)y_i[n] = (1+(n-1)k)y_i[n-1] + (1+(n-1)k)\alpha_i[n] - (1+(n-2)k)\alpha_i[n-1]
\]
Divide both sides by $1+(n-1)k$, we get
\[
y_i[n] = y_i[n-1] + \alpha_i[n] - \frac {1+(n-2)k} {1+(n-1)k} \alpha_i[n-1]
\]
By iteration, we have
\[
\begin{split}
y_i[n] &= y_i[n-2] + \alpha_i[n-1] - \frac {1+(n-3)k} {1+(n-2)k} \alpha_i[n-2] + \alpha_i[n] - \frac {1+(n-2)k} {1+(n-1)k} \alpha_i[n-1]\\
&= y_i[n-2] + \alpha_i[n] + \frac k {1+(n-1)k} \alpha_i[n-1] - \frac {1+(n-3)k} {1+(n-2)k} \alpha_i[n-2]\\
&= y_i[n-3] + \alpha_i[n-2] - \frac {1+(n-4)k} {1+(n-3)k} \alpha_i[n-3] + \alpha_i[n] + \frac k {1+(n-1)k} \alpha_i[n-1] - \frac {1+(n-3)k} {1+(n-2)k} \alpha_i[n-2]\\
&= y_i[n-3] + \alpha_i[n] + \frac k {1+(n-1)k} \alpha_i[n-1] + \frac k {1+(n-2)k} \alpha_i[n-2] - \frac {1+(n-4)k} {1+(n-3)k} \alpha_i[n-3]\\
&= \dots\\
&= y_i[1] + \alpha_i[n] + k\sum_{j=2}^{n-1} \frac 1 {1+jk} \alpha_i[j] - \frac 1 {1+k} \alpha_i[1]\\
&= N_{\boldsymbol x;v_i} + \alpha_i[1] + \alpha_i[n] + k\sum_{j=2}^{n-1} \frac 1 {1+jk} \alpha_i[j] - \frac 1 {1+k} \alpha_i[1] \quad (\text{Substitute the initial value }y_i[1] = N_{\boldsymbol x;v_i} + \alpha_i[1])\\
&= N_{\boldsymbol x;v_i} + \alpha_i[n] + k\sum_{j=2}^{n-1} \frac 1 {1+jk} \alpha_i[j] + \frac k {1+k} \alpha_i[1]\\
&= N_{\boldsymbol x;v_i} + \alpha_i[n] + k\sum_{j=1}^{n-1} \frac 1 {1+jk} \alpha_i[j]\\
\end{split}
\]
Therefore, we conjecture that \cref{subsample-equation} has a solution
\[
\begin{cases}
y_i[n] &=~~N_{\boldsymbol x;v_i} + \alpha_i[n] + k\sum_{j=1}^{n-1} \frac 1 {1+jk} \alpha_i[j]\\
y[n] &=~~N_{\boldsymbol x} + \alpha[n] + k\sum_{j=1}^{n-1} \frac 1 {1+jk} \alpha[j]\\
\end{cases}
\]
The following section proves that it is a solution to \cref{subsample-equation}.

\subsection{Proof that \cref{subsample-solution} is a solution to the recurrence \cref{subsample-equation}}
\label{sec: subsample-proof}

Substitute $n = 1$ into
\[
\begin{cases}
y_i[n] &=~~N_{\boldsymbol x;v_i} + \alpha_i[n] + k\sum_{j=1}^{n-1} \frac 1 {1+jk} \alpha_i[j]\\
y[n] &=~~N_{\boldsymbol x} + \alpha[n] + k\sum_{j=1}^{n-1} \frac 1 {1+jk} \alpha[j]\\
\end{cases},
\]
we get
\[
\begin{cases}
y_i[1] &=~~N_{\boldsymbol x;v_i} + \alpha_i[1]\\
y[1] &=~~N_{\boldsymbol x} + \alpha[1]\\
\end{cases},
\]
which satisfies the initial condition
\[
\hat p_1(v_i|\boldsymbol x) = \frac {y_i[1]} {y[1]} = \frac {N_{\boldsymbol x;v_i} + \alpha_i[1]} {N_{\boldsymbol x} + \alpha[1]}.
\]
Substitute
\[
\begin{cases}
y_i[j] &=~~N_{\boldsymbol x;v_i} + \alpha_i[j] + k\sum_{m=1}^{j-1} \frac 1 {1+mk} \alpha_i[m]\\
y[j] &=~~N_{\boldsymbol x} + \alpha[j] + k\sum_{m=1}^{j-1} \frac 1 {1+mk} \alpha[m]\\
\end{cases}
\]
into the right hand side of
\[
\hat p_n(v_i|\boldsymbol x) = \frac {\frac {N_{\boldsymbol x;v_i} + k\sum_{j=1}^{n-1} y_i[j]} {1+(n-1)k} + \alpha_i[n]} {\frac {N_{\boldsymbol x} + k\sum_{j=1}^{n-1} y[j]} {1+(n-1)k} + \alpha[n]},
\]
we get
\[
\begin{split}
numerator &= \frac {N_{\boldsymbol x;v_i} + k\sum_{j=1}^{n-1} (N_{\boldsymbol x;v_i} + \alpha_i[j] + k\sum_{m=1}^{j-1} \frac 1 {1+mk} \alpha_i[m])} {1+(n-1)k} + \alpha_i[n]\\
\\
&= \frac {N_{\boldsymbol x;v_i} + (n-1)kN_{\boldsymbol x;v_i} + k\sum_{j=1}^{n-1} (\alpha_i[j] + k\sum_{m=1}^{j-1} \frac 1 {1+mk} \alpha_i[m])} {1+(n-1)k} + \alpha_i[n]\\
\\
&= \frac {(1+(n-1)k)N_{\boldsymbol x;v_i} + k\sum_{j=1}^{n-1} (\alpha_i[j] + k\sum_{m=1}^{j-1} \frac 1 {1+mk} \alpha_i[m])} {1+(n-1)k} + \alpha_i[n]\\
\\
&= N_{\boldsymbol x;v_i} + \alpha_i[n] + k\frac {\sum_{j=1}^{n-1} (\alpha_i[j] + k\sum_{m=1}^{j-1} \frac 1 {1+mk} \alpha_i[m])} {1+(n-1)k}\\
\\
&= N_{\boldsymbol x;v_i} + \alpha_i[n] + k\frac {\sum_{j=1}^{n-1} \alpha_i[j] + \sum_{j=1}^{n-1} (k\sum_{m=1}^{j-1} \frac 1 {1+mk} \alpha_i[m])} {1+(n-1)k}\\
\\
&= N_{\boldsymbol x;v_i} + \alpha_i[n] + k\frac {\sum_{j=1}^{n-1} \alpha_i[j] + k\sum_{j=1}^{n-1}\sum_{m=1}^{j-1} \frac 1 {1+mk} \alpha_i[m]} {1+(n-1)k}\\
\\
&= N_{\boldsymbol x;v_i} + \alpha_i[n] + k\frac {\sum_{j=1}^{n-1} \alpha_i[j] + k\sum_{m=1}^{n-2} \frac {n-1-m} {1+mk} \alpha_i[m]} {1+(n-1)k}\\
\\
&= N_{\boldsymbol x;v_i} + \alpha_i[n] + k\frac {\alpha_i[n-1] + \sum_{j=1}^{n-2} \alpha_i[j] + k\sum_{j=1}^{n-2} \frac {n-1-j} {1+jk} \alpha_i[j]} {1+(n-1)k}\\
\\
&= N_{\boldsymbol x;v_i} + \alpha_i[n] + k\frac {\alpha_i[n-1] + \sum_{j=1}^{n-2} (1 + k\frac {n-1-j} {1+jk}) \alpha_i[j]} {1+(n-1)k}\\
\\
&= N_{\boldsymbol x;v_i} + \alpha_i[n] + k(\frac {\alpha_i[n-1]} {1+(n-1)k} + \frac {\sum_{j=1}^{n-2} \frac {1+jk+kn-k-jk} {1+jk} \alpha_i[j]} {1+(n-1)k})\\
\\
&= N_{\boldsymbol x;v_i} + \alpha_i[n] + k(\frac {\alpha_i[n-1]} {1+(n-1)k} + \frac {\sum_{j=1}^{n-2} \frac {1+(n-1)k} {1+jk} \alpha_i[j]} {1+(n-1)k})\\
\\
&= N_{\boldsymbol x;v_i} + \alpha_i[n] + k(\frac {\alpha_i[n-1]} {1+(n-1)k} + \sum_{j=1}^{n-2} \frac 1 {1+jk} \alpha_i[j])\\
\\
&= N_{\boldsymbol x;v_i} + \alpha_i[n] + k\sum_{j=1}^{n-1} \frac 1 {1+jk} \alpha_i[j]\\
\\
&= y_i[n],\\
\end{split}
\]
\[
\begin{split}
denominator &= \frac {N_{\boldsymbol x} + k\sum_{j=1}^{n-1} (N_{\boldsymbol x} + \alpha[j] + k\sum_{m=1}^{j-1} \frac 1 {1+mk} \alpha[m])} {1+(n-1)k} + \alpha[n]\\
\\
&= N_{\boldsymbol x} + \alpha[n] + k\frac {\sum_{j=1}^{n-1} \alpha[j] + k\sum_{j=1}^{n-1}\sum_{m=1}^{j-1} \frac 1 {1+mk} \alpha[m]} {1+(n-1)k}\\
\\
&= N_{\boldsymbol x} + \alpha[n] + k\frac {\sum_{j=1}^{n-1} \alpha[j] + k\sum_{m=1}^{n-2} \frac {n-1-m} {1+mk} \alpha[m]} {1+(n-1)k}\\
\\
&= N_{\boldsymbol x} + \alpha[n] + k(\frac {\alpha[n-1]} {1+(n-1)k} + \frac {\sum_{j=1}^{n-2} \frac {1+(n-1)k} {1+jk} \alpha[j]} {1+(n-1)k})\\
\\
&= N_{\boldsymbol x} + \alpha[n] + k\sum_{j=1}^{n-1} \frac 1 {1+jk} \alpha[j]\\
\\
&= y[n],\\
\\
\\
\frac {numerator} {denominator} &= \frac {y_i[n]} {y[n]} = \hat p_n(v_i|\boldsymbol x).\\
\\
\end{split}
\]
This shows that the obtained solution satisfies the original recurrence equation.

% \[
% y_i[n] = N_{\boldsymbol x;v_i} + 1 + \sum_{j=1}^{n-1} \frac 1 {1+j} = N_{\boldsymbol x;v_i} +\sum_{j=1}^n \frac 1 j
% \]

\section{Experimental details}
\label{sec: methods}

\subsection{Recursive training}
\label{sec: recursive}

We train the LM at generation $n$ exclusively on the text generated by generation $n-1$. In this \lcnamecref{sec: recursive}, we begin by introducing the formal definition of this recursive process. Then, we elaborate on the details, including hyperparameters used in our training.

\subsubsection{Formal definition}

We define our recursive training process by a first-order Markov process:
\[
D \to \boldsymbol\theta[1] \to \boldsymbol\theta[2] \to \boldsymbol\theta[3] \dots,
\]
where $\boldsymbol\theta[n]$ is the vector of all weights of generation $n$, subject to the first-order Markov assumption: given the values of all weights of generation $n$, then the probability of the weights of generation $n+1$ gets a certain set of values, is conditionally independent of the weights of generation $n-1$ and before, including the initial text corpus, $D$, that is
\[
P\big(\boldsymbol\theta[n+1]~|~\boldsymbol\theta[n],~\boldsymbol\theta[n-1],\dots,\boldsymbol\theta[1],~D\big) = P\big(\boldsymbol\theta[n+1]~|~\boldsymbol\theta[n]\big).
\]

\subsubsection{Details and hyperparameters}

The transition from $\boldsymbol\theta[n]$ to $\boldsymbol\theta[n+1]$ is defined by the following stochastic algorithm:

Randomly initializes the weights of generation $n+1$; Update the weights by generative pre-training the LM, that is, gradient descent of the cross entropy loss between the target token and the LM's prediction given a preceding token sequence; The target token is generated by the generation $n$ by faithfully (\ie with \verb+temperature=1+) sampling from its output distribution given that preceding token sequence. \cref{a} is the pseudocode of this algorithm.

With some hyperparameters placed at their proper place, a minimized pseudocode snippet that can be used to reproduce the state transition in the above Markov process is \cref{a}. The values of hyperparameters follow what \citet{TinyStories} used to train their 33M LM (\url{https://huggingface.co/roneneldan/TinyStories-33M}) unless specified. We provide detailed comment for each operation in the pseudocode. We use function and class names resembling those in the \textit{Transformers} library \citep{Transformers}.

\begin{algorithm}
\caption{State transition from $\boldsymbol\theta[n]$ to $\boldsymbol\theta[n+1]$ in the Markov process}
\label{a}
\begin{algorithmic}[1]
\Require $m$, an instance of \textsc{CausalLanguageModel}. $\boldsymbol\theta[n]$ is the vector of its weights before executing this algorithm.
\Require \textsc{Num-Training-Tokens}, a hyperparameter. We set this to 600,000,000.
\Require \textsc{Max-Context-Length}, a hyperparameter. We set this to 512.
\Ensure $m'$, a new instance of \textsc{CausalLanguageModel}. $\boldsymbol\theta[n+1]$ is the vector of its weights after executing this algorithm.
\LComment{Get the old LM's config to ensure that the new LM's architecture, parameter size, and vocabulary are the same as the old LM. The new LM's weights are randomly initialized.}
\State $m' \gets \Call{CausalLanguageModel.from-config}{m\textsc{.config}}$
\State $optimizer \gets \Call{AdamW}{m'\textsc{.parameters},~\textsc{Learning-Rate=}5\times10^{-4},\\\textsc{Betas=}(0.9,0.95),~\textsc{Weight-Decay=}0.1}$
\State $context \gets$ a queue of tokens, initially empty
\For{$i = 1, \dots, \textsc{Num-Training-Tokens}$}
    \State $\boldsymbol p \gets m(context)$ \label{predict}
    \State $\boldsymbol q \gets m'(context)$
    \State $x \gets \Call{Sample}{\boldsymbol p}$ \Comment{Sample a single token from $\boldsymbol p$.} \label{sample}
    \State $loss \gets \Call{Cross-Entropy}{\Call{One-Hot}{x},~\boldsymbol q}$
    \Comment{\Call{One-Hot}{x} is a distribution on the LM's vocabulary, with $x$ as the only token with non-zero probability.}
    \State $loss\textsc{.backward}()$ \Comment{Use backpropagation, calculate the gradient\dots}
    \State $optimizer\textsc{.step}()$ \Comment{\dots to update the weights of the new LM.}
    \State $context\textsc{.enqueue}(x)$ \label{enqueue}
    \If{$context\textsc{.length} > \textsc{Max-Context-Length}$}
        \State $context\textsc{.dequeue}()$
    \EndIf
\EndFor
\end{algorithmic}
\end{algorithm}

In our practice, we batch the tokens and accumulate the gradients for several update steps before performing a backward/update pass to speed up training. The total effective training batch size is 320. We enabled fp16 16-bit (mixed) precision training instead of 32-bit training. Lines~\ref{predict}, \ref{sample}, and~\ref{enqueue} of \cref{a} are executed separately to generate a training corpus before we train a new LM. We generate the corpus in parallel, with 200 as the batch size. Therefore, it contains 1,000,000 independent sequences. The length per sequence is 600 tokens. The new LM is then trained on this corpus for 3 epochs. We hold out 5\% of texts from \TN~as a validation set to determine the number of epochs by early stopping during the training of the first LM. We reuse this number of epochs in subsequent training.

We utilized a single NVIDIA RTX™ A6000 GPU as our computing device. Training each generation of the 33M LM required no more than one day. With a total of 40 generations trained, the entire process was completed within a month.

\begin{table}
    \caption{\textbf{Excerpts from generated texts of different generations of 33M LMs.} Texts generated by generation $n-1$ form the training corpus of generation $n$.}
    \label{tab: excerpts}
    \centering
    \begin{small}
    \begin{tabular}{@{}lp{0.8\textwidth}@{}}
        \toprule
        Generation & Sample\\
        \midrule
        1 & there was a little boy called Sam. He was three years old and loved playing outside.\\
        1 & there lived a big boy. His name was Joe. Joe was very proud of his cool cap. He kept it close and never wanted anyone else to touch it.\\
        6 & Mary and Charlie were best friends. They played together and were always so excited to go for a ride.\\
        6 & It was going on a special day. A boy was walking at a shop. He was excited because he had never seen a big bag of treats on this menu.\\
        11 & Ellie Ellie was feeling excited. She was ready to go on an adventure.\\
        11 & Mom side, a little girl wanted to go and see around the house. She asked her mother if she could go.\\
        16 & did day one day, something went by her pocket. She put on a ribbon and decided to let her pass in the garden.\\
        16 & "What do you see, Mommy?" she asked asked, pointing to the fancy dishes.\\
        21 & Mia in the garden and an modest dog named Tim, who always looked for his might. He liked to play with his owner, and who went off the house.\\
        21 & All the animals were happy in the squ sure together. So one day, the animals played and chased especially another friend.\\
        26 & Ellie herea and her mom decided to play together. They drove in the car. Dad said they are been friends.\\
        26 & bar bar bar bar bar bar bar bar bar bar bar bar bar bar bar bar bar bar bar bar bar\\
        31 & did day one day, something very green-f anyone could always know becausealia at action was before\\
        31 & Lib after touch wherever looked troubled, kind much fun they both both weighed bright butterfly faces nurses zipping z z z z z z\\
        36 & frog frog frog frog frog frog frog frog frog frog frog frog frog frog frog frog frog frog frog frog frog frog\\
        36 & P Pink Spring Spring Spring Spring Spring spring Ada, mun-of squ sure up spring. Every morning he was made fun- leaping especially another spring morning p guilty.\\
        \bottomrule
    \end{tabular}
    \end{small}
\end{table}

\subsection{LM architecture}
\label{sec: architecture}

A 1M LM has 8 layers and a hidden size 64, with 16 attention heads. It supports a maximum position embedding of 2048 and has a vocabulary size of 50257, utilizing global and local attention mechanisms. In contrast, a 33M LM features 4 layers and a larger hidden size of 768 with 16 attention heads. It shares the same maximum position embedding of 2048 and vocabulary size of 50257, incorporating the same attention mechanisms as the 1M LM. These configurations are the configurations of \TN-1M and \TN-33M made available by \citet{TinyStories} through Hugging Face.

\subsection{Model perplexity}
\label{sec: perplexity}

We calculate the model perplexity to evaluate how well an LM predicts the next token in the sequence, with lower perplexity indicating better performance. To calculate the perplexity, we must first calculate the \emph{token loss} and then the \emph{validation loss}.

\paragraph{Token loss of a token in the validation set} The validation set of \TN~consist of short stories. Each story is a sequence of tokens $\boldsymbol x$. Let $x_t$ denote the $t$-th token of $\boldsymbol x$. Let $context(x_t)$ denote the tokens before $x_t$ within the same sequence. Specifically, we accommodate context window constraints to align with the model’s architectural limits. When $t \leq 512$, the context is $\boldsymbol x_{<t}$; however, when $t > 512$, the model only considers the previous 512 tokens due to its context window limitation, which adjusts the context to $\boldsymbol x_{t-512:t}$. We define $\hat p(x_t|context(x_t))$ as given $context(x_t)$, the LM predict that the next token is $x_t$ with probability $\hat p(x_t|context(x_t))$. For each sequence $\boldsymbol x$ in the validation set, the token loss of its $t$-th token is $-\log \hat p(x_t|context(x_t))$.

\paragraph{Validation loss $L$} is the arithmetic average of the token losses of all tokens in the validation set.

Finally, we calculate the model perplexity by $e^L$.

\end{document}